%% file: main.tex
\pdfoutput=1

\documentclass[11pt]{article}

\usepackage[preprint]{acl}

\usepackage{times}
\usepackage{latexsym}

\usepackage[T1]{fontenc}

\usepackage[utf8]{inputenc}

\usepackage{microtype}

\usepackage{inconsolata}

\usepackage{graphicx}
\usepackage{booktabs}
\usepackage{url}
\usepackage{amsmath}
\usepackage{amssymb}
\usepackage{colortbl}
\usepackage{multirow}
\usepackage{makecell}
\PassOptionsToPackage{prologue,dvipsnames}{xcolor}
\usepackage[dvipsnames]{xcolor}
\definecolor{mylightred}{HTML}{FFE9E8}
\definecolor{mydrakgreen}{HTML}{27967e}
\usepackage{color}
\usepackage{ulem}
\usepackage{paralist}
\usepackage{pifont}
\usepackage{marvosym}

\usepackage[most]{tcolorbox}
\usepackage{csquotes}
\usepackage{anyfontsize}
\usepackage{comment}
\usepackage{float}
\usepackage{cuted}
\usepackage{tikz}
\usepackage{listings,multicol}
\usepackage[
    framemethod=tikz,
    skipbelow=\topskip,
    skipabove=\topskip
]{mdframed}
\newtcolorbox[list inside=prompt,auto counter,number within=section]{prompt}[1][]{
    colbacktitle=black!60,
    coltitle=white,
    fontupper=\footnotesize,
    boxsep=5pt,
    left=0pt,
    right=0pt,
    top=0pt,
    bottom=0pt,
    boxrule=1pt,
    #1,
}

\title{Towards Omni-RAG: Comprehensive Retrieval-Augmented Generation \\ for Large Language Models in Medical Applications}

\author{
    Zhe~Chen$^{1,3}$,
    Yusheng~Liao$^{1,3}$,
    Shuyang~Jiang$^{2,3}$,
    Pingjie~Wang$^{1,3}$
    \\
    \bf
    Yiqiu~Guo$^{2,3}$,
    Yanfeng~Wang$^{1,3}$,
    Yu~Wang$^{1,3}$\textsuperscript{\Letter} \vspace{0.6mm}
    \\ 
    \begin{tabular}{c} 
    $^1$Shanghai Jiao Tong University ~~~
    $^2$Fudan University \\
    $^3$Shanghai Artificial Intelligence Laboratory \vspace{0.6mm}\\
    \end{tabular}
    \\ 
    \small
    \begin{tabular}{c}
    \texttt{\{chenzhe2018,liao20160907,pingjiewang,wangyanfeng622,yuwangsjtu\}@sjtu.edu.cn} \\
    \texttt{\{shuyangjiang23,yqguo22\}@m.fudan.edu.cn} \\
    \end{tabular}
}

\begin{document}
\maketitle

\renewcommand{\thefootnote}{}
\footnotetext{\Letter: Corresponding author.}
\renewcommand{\thefootnote}{\arabic{footnote}} 

\input{00_abstract}
\input{01_introduction}

\input{02_related_work}

\input{03_the_xxx_data}
\input{04_method}
\input{05_experiments}
\input{06_conclusion}
\input{07_limit_ethi}

\section*{Acknowledgements}
This work was supported by the National Key R\&D Program of China (No. 2022ZD0162101) and STCSM (No. 22DZ2229005).

\bibliography{medrag}
\appendix
\input{08_appendix}

\end{document}

%% file: 00_abstract.tex
\begin{abstract}

Large language models hold promise for addressing medical challenges, such as medical diagnosis reasoning, research knowledge acquisition, clinical decision-making, and consumer health inquiry support. However, they often generate hallucinations due to limited medical knowledge. Incorporating external knowledge is therefore critical, which necessitates multi-source knowledge acquisition. We address this challenge by framing it as a source planning problem, which is to formulate context-appropriate queries tailored to the attributes of diverse sources. Existing approaches either overlook source planning or fail to achieve it effectively due to misalignment between the model's expectation of the sources and their actual content. To bridge this gap, we present MedOmniKB, a repository comprising multigenre and multi-structured medical knowledge sources. Leveraging these sources, we propose the Source Planning Optimisation method, which enhances multi-source utilisation. Our approach involves enabling an expert model to explore and evaluate potential plans while training a smaller model to learn source alignment. Experimental results demonstrate that our method substantially improves multi-source planning performance, enabling the optimised small model to achieve state-of-the-art results in leveraging diverse medical knowledge sources\footnote{Project website: \url{https://github.com/Jack-ZC8/Omni-RAG-Medical}}.

\end{abstract}

%% file: 01_introduction.tex
\section{Introduction}

Large language models (LLMs) have demonstrated impressive language capabilities \cite{Zhu2023, Zhou2023, Chen2023a, Singhal2023, Nori2023, Zhu2025, Chen2024a, Saab2024}. However, these models can produce factually incorrect responses—often termed “hallucinations”—when their internal knowledge is insufficient \cite{Ji2023, Jiang2025}. Such inaccuracies pose significant challenges for medical applications, where correctness and trustworthiness are paramount. To mitigate this issue, recent work has explored medical Retrieval-Augmented Generation (RAG) techniques \cite{Xiong2024, Yang2024, Jeong2024, Xiong2024a, Xu2024a}, which enhance the accuracy and transparency of model responses \cite{Zhu2024}.

\begin{figure}[t]
    \centering
    \includegraphics[width=0.9\linewidth]{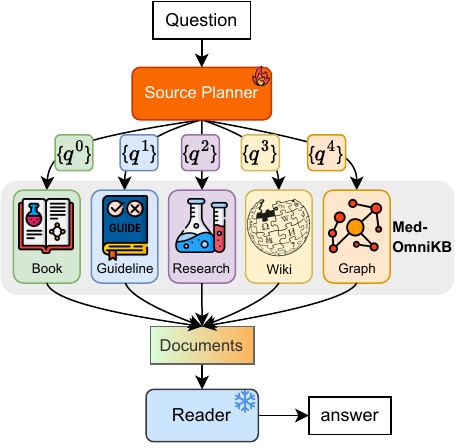}
    \caption{Diagram of source planning in medical scenarios.}
    \label{fig:source_planning_problem}
    \vspace{-0.4cm}
\end{figure}

These challenges span various domains, including medical reasoning for diagnoses and treatments, knowledge acquisition from advanced research, clinical decision-making using patient data, and comprehensive responses to health-related inquiries. Addressing these complex healthcare questions requires the integration of diverse knowledge sources~\cite{Xiong2024, Xu2024, Corbeil2024}. Existing methods typically treat all sources uniformly, using the original question to retrieve without tailoring the search strategy to different sources \cite{Xiong2024, Xu2024, Neupane2024, Jeong2024}. Although subsequent approaches have introduced prompting strategies to guide LLMs in leveraging retrieval sources \cite{Ma2023, Jiang2024, Li2024a}, these still fail to construct proper queries for each source. The restricted information presented in the source description prompts causes the misalignment between what the model expects from the sources and what the sources actually contain. Similarly, recent reflection methods \cite{Shinn2023, Hu2024, Liao2024, Zhao2024}, which iteratively retrieve and self-improve queries, also struggle due to the limited self-reflective capacity when facing multiple sources. In essence, current works cannot construct appropriate queries for each knowledge source based on its unique attributes, which is referred to as the “source planning” problem in our work (see Figure~\ref{fig:source_planning_problem}).

A key obstacle in studying the problem is the lack of sufficiently broad and diverse medical knowledge bases. To address this, we introduce MedOmniKB, a more comprehensive and varied knowledge repository than previously available resources \cite{Xiong2024, Corbeil2024, Xu2024, Lin2024a}. Five representative sources—“Book,” “Guideline,” “Research,” “Wiki,” and “Graph”—offer both depth and breadth of information. Its diversity in categories allows for a more meaningful exploration of source planning.

Leveraging MedOmniKB, we propose a Source Planning Optimisation (SPO), a novel paradigm for knowledge integration. Our approach proceeds as follows: First, we use an expert LLM to perform planning exploration for each question, generating candidate sub-plans. Next, we evaluate whether the documents retrieved by these sub-plans support the correct answer, thereby creating positive and negative plan sets. Finally, we employ a smaller language model to undergo supervised fine-tuning (SFT) on the positive plans and then apply Direct Preference Optimisation (DPO) \cite{Rafailov2023} to further align the model with the knowledge sources. Extensive experiments show that SPO substantially boosts multi-source planning capability compared to existing techniques. Notably, our optimised small model outperforms substantially larger models (with 10 times the parameters) in retrieval planning. Furthermore, SPO demonstrates robustness under various training conditions and achieves a high utilisation rate of training data.

In summary, our contributions are three-fold:
\begin{compactitem}
\item We identify the challenge of multi-source planning in medicine and introduce MedOmniKB as a foundation for research in this area. It addresses the critical gap in richly diverse, large-scale medical knowledge resources.

\item Based on MedOmniKB, we propose the SPO approach, a novel paradigm for knowledge integration, empowering language models to adapt retrieval strategies for diverse sources.

\item Extensive experiments confirm that the optimised model achieves more efficient source planning than existing methods, offering insights into multi-source retrieval strategies and expanding the applications of large language models in the medical field.
\end{compactitem}

%% file: 02_related_work.tex
\section{Related Work}

\subsection{Medical Retrieval-augmented Generation}

RAG has been successfully applied to a broad spectrum of medical tasks, including clinical decision-making~\cite{Shi2023, Thompson2023}, clinical prediction~\cite{Ye2021, Naik2022, Xu2024}, and medical question-answering (QA)~\cite{Xiong2024, Jeong2024, Wang2024b, Li2024a}. Among these applications, medical QA presents significant challenges due to its demand for extensive and precise knowledge integration. While existing studies have developed multi-content retrieval engines for medical information~\cite{Xiong2024, Corbeil2024, Xu2024}, our approach offers a distinct advantage by utilizing a substantially larger knowledge base that incorporates structured knowledge graphs. As illustrated in Table~\ref{tab:compare_kb}, our knowledge base not only surpasses others in size but also enhances data organization and retrieval accuracy through structured representations.

\subsection{Query Construction}
Constructing queries is crucial for retrieval augmented generation. Some works directly use large language models to enhance the query~\cite{Ma2023, Wang2023, Wang2024b, Wu2024a, Wang2024a, Chen2024}, and others rely on multiple retrievals and reflection to improve the query quality~\cite{Shinn2023, Hu2024, Zhao2024, Liao2024}. They all struggle to construct proper queries because they either lack the perception of sources or have only limited self-reflective capacity.

There are also studies focused on constructing query-related training data. For example, \citet{Ma2023} evaluate query value based on the performance of the downstream task, which may be influenced by the intrinsic knowledge of LLMs. \citet{Mao2024} employ rerank scores directly, which provide a fast approach but are primarily limited to relevance ranking. \citet{Chan2024} generate three types of enhanced queries using ChatGPT as training data; however, their work does not explicitly assess the quality of these enhanced queries. \citet{Yoon2024, Wang2024} incorporate retriever feedback, which can be effective but often relies on gold document annotations, which may not always be available in general scenarios. In contrast, we adopt the LLM-as-a-judge paradigm~\cite{Li2024} to obtain high-quality training data through LLM-based evaluation. Additionally, while \citet{Wang2023b, Wang2024c} focus primarily on source selection, they place less emphasis on query reformulation. This highlights the need for further exploration of effective retrieval source planning.

%% file: 03_the_xxx_data.tex
\section{MedOmniKB} \label{sec:medkb}

\begin{table}[t]
    \resizebox{\linewidth}{!}{
    \centering
    \begin{tabular}{lccccccccc} 
        \toprule[2pt]
        \bf Database & \bf \#Book & \bf \#Guideline & \bf \#Research & \bf \#Wiki & \bf \#Graph \\
        \midrule
        MedCorp~\citeyearpar{Xiong2024} & 9.6k & - & 23.9M & 6.5M & \ding{55} \\
        ClinicalCorp~\citeyearpar{Corbeil2024} & 9.4k & 46.1k & 150.4k & - & \ding{55} \\
        Self-BioRAG~\citeyearpar{Jeong2024} & 18 & 35.7k & 37.5M & - & \ding{55} \\ 
        RAM-EHR~\citeyearpar{Xu2024} & - & - & 230k & 150k & \ding{52} \\
        BioKGBench~\citeyearpar{Lin2024a} & - & - & 5.7k & - & \ding{52} \\
        \midrule
        MedOmniKB & 27.7k & 45.7k & 25.3M & 6.4M & \ding{52}\\
        \bottomrule[2pt]
    \end{tabular}
    }
    \caption{Comparison of MedOmniKB with existing medical retrieval knowledge bases in terms of \#Docs. ``-'' denotes the type doesn't exist.}
    \label{tab:compare_kb}
    \vspace{-0.2cm}
\end{table}

This section introduces MedOmniKB, a comprehensive multigenre, multistructured medical knowledge base. We investigate the knowledge sources required for existing medical knowledge-intensive problems and correspondingly organise a knowledge base containing five sources: ``Book'', ``Guideline'', ``Research'', ``Wiki'' and ``Graph''. The richness and diversity of categories establish the basis for our study of medical source planning. The knowledge base is intended to be used only for research purposes. We did not anonymise them, as they are public medical corpora.

\subsection{Unstructured Source}

\paragraph{Book.}
Medical textbooks encompass foundational medical knowledge, which is crucial for understanding the field of medicine. Following \citet{Jin2020, Wu2024}, 18,182 PDFs are gathered from online libraries and reputable publishers, whose categories cover medicine, surgery, imaging, etc. Their details are in Appendix~\ref{appendix:db}. These PDFs were de-duplicated and filtered (URLs, references, citations, etc.) to get documents. We also include StatPearls and Textbooks proposed in MedRAG~\cite{Xiong2024}.
\paragraph{Guideline.}
Clinical practice guidelines help healthcare practitioners and patients make decisions about diagnosis and treatment. We reuse the guideline data from \cite{Chen2023}, and for the non-redistributable parts, we crawl webpages using the provided scripts. In the end, we get 45,679 articles from 13 guideline sources.
\paragraph{Research.}
Research articles offer insights and findings from cutting-edge research, thereby providing a solid theoretical basis for clinical practice and public health decision-making. We download the 2024 PubMed baseline\footnote{\url{https://ftp.ncbi.nlm.nih.gov/pubmed/baseline}}, which is a complete snapshot of PubMed data and then extract the valid titles and abstracts.
\paragraph{Wiki.}
Wikipedia acts as a valuable resource by providing knowledge in general domains. We obtain the processed English data from Huggingface\footnote{\url{https://huggingface.co/datasets/wikimedia/wikipedia}}.

For retrieval, we process these texts into chunks of no more than 1000 characters, followed by encoding them as vectors using the MedCPT-article-encoder model~\cite{Jin2023}. We adopt the Qdrant library\footnote{\url{https://qdrant.tech}} for fast dense vector searching.

\subsection{Structured Source}

\begin{table}[t] \small
    \centering
    \begin{tabular}{lccccccccc} 
        \toprule[1pt]
        \textbf{Source} & \bf \#Docs & \bf \#Chunks & \bf \#Words/Chunk\\
        \midrule
        Book & 27.7k & 13.1M & 150.1 \\
        Guideline & 45.7k & 647.7k & 106.7 \\
        Research & 25.3M & 48.0M & 128.7 \\
        Wiki & 6.4M & 29.7M & 112.1 \\
        \midrule
    \end{tabular}
    \begin{tabular}{lccccccccc} 
         & \bf \#Concepts & \bf \#Definitions & \bf \#Relations\\
        \midrule
        Graph & 1.7M & 317.9k & 2.9M \\
        \bottomrule[1pt]
    \end{tabular}
    \caption{Statistics of MedOmniKB.}
    \label{tab:stat}
\end{table}

\begin{figure*}[htbp]
    \centering
    \includegraphics[width=1\linewidth]{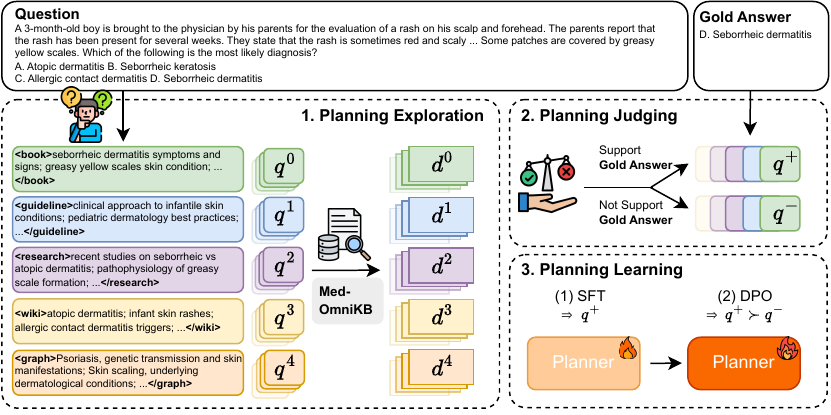}
    \caption{Framework of our proposed SPO approach. First, an expert LLM performs plan exploration for each source, generating multiple queries. Then the expert LLM determines whether the documents retrieved by each query support the gold answer or not. The final positive and negative sets are utilised for planning learning.}
    \label{fig:spo}
\end{figure*}

\paragraph{Graph.}
Structured graphs provide clear definitions of concepts and illustrate relationships between them, potentially improving evidence-based decision-making in healthcare. First, the Unified Medical Language System (UMLS)~\cite{Bodenreider2004} Metathesaurus Full Subset is downloaded and cleaned. Then, the description, indications, pharmacodynamics, absorption, and drug interactions of drugs stored in DrugBank~\cite{Wishart2008} are added to the definition of the corresponding nodes. The combined definitions and relationships are stored in SQLite\footnote{\url{https://www.sqlite.org}}, avoiding the network and latency issues of the online UMLS API. 

For a given concept, we will obtain its definition and all its one-hop relationships. Next, reranking will be applied to filter through the many relationships (possibly thousands) following \citet{Yang2023, Yang2024}.

%% file: 04_method.tex
\section{Method}

\subsection{Problem Formulation}

We provide a general formulation of our medical source planning problem. For a medical question $x$, multiple knowledge sources $K = \{ K^i \mid i = 1, 2, \ldots, N_{K} \}$ will be retrieved to supply medical knowledge, where $K^i$ denotes the $i_{th}$ source and $N_{K}$ denotes the number of sources. Specifically, a planner model $\mathcal{M}_{\theta}$ will construct the source plan $P = \{ (i, j, q_{j}^{i}) \mid i = 1, 2, ..., N_{K}, \; j = 1, 2, ..., N^{i}_{q} \}$ for the question $x$, where $q_{j}^{i}$ denotes the $j_{th}$ query in the $i_{th}$ source, and $N^{i}_{q}$ represents the number of queries for the $i_{th}$ source. We then obtain the retrieved document $D=\{d_{j}^{i} \mid i = 1, 2, ..., N_{K}, \; j = 1, 2, ..., N^{i}_{q}\}$, where $d_{j}^{i}$ represents the top-$k$ documents retrieved by the query $q_{j}^{i}$. Then a language model Reader will read $D$ and answer question $x$, resulting in an answer $y$. Our task is to construct the optimal $P$ so that $D$ includes as many documents supporting the gold answer as possible. Additionally, the number of queries per source $N^{i}_{q}$ will be limited to fewer than 4 due to the context length restriction of the Reader.

\subsection{Source Planning Optimisation}

In this section, we introduce our Source Planning Optimisation (SPO) method to empower the planner $\mathcal{M}_{\theta}$ to retrieve valuable information from multiple healthcare knowledge sources efficiently. Our method comprises three steps: Planning Exploration (Section~\ref{sec:plan_exp}), Planning Judging (Section~\ref{sec:plan_ver}) and Planning Learning (Section~\ref{sec:plan_learn}).

\subsubsection{Planning Exploration}\label{sec:plan_exp}

We begin with planning exploration to identify potential multi-source planning strategies. This process is performed on a training set consisting of comprehensive, knowledge-intensive medical problems. In our approach, we prompt an expert LLM, Qwen2.5-72B-Instruct-AWQ~\cite{QwenTeam2024}, to generate multiple queries for each source. The exploration prompt is guided by two principles: diversity within a single source and alignment with the characteristics of different sources\footnote{The exploration prompt is shown in Prompt~\ref{prompt:plan_explor}.}. The diversity principle encourages a wide range of queries from the same source, promoting varied perspectives and minimizing redundancy. The alignment principle ensures that queries are tailored to fit the unique properties of each source. Experimental results on query diversity are discussed in Appendix~\ref{appendix:diversity}, and the number of queries per source is fixed at six.

Subsequently, we use the queries to retrieve their corresponding sources and gather the top-$k$ documents for each query. The superscripts ``$+$'' and ``$-$'' represent positive and negative ones, respectively.

\subsubsection{Planning Judging}\label{sec:plan_ver}

Inspired by the recent emergence of LLM-as-a-judge~\cite{Li2024}, we prompt the superior LLM Qwen2.5-72B-Instruct-AWQ to judge whether the documents retrieved by the query support the gold answer\footnote{The judging prompt is shown in Prompt~\ref{prompt:plan_judge}.}. Each $q^i_{j}$ will be categorized as ``positive'' or ``negative'', which can be formulated as:
\begin{equation} 
q^i_{j} = 
\begin{cases} 
q^{i,+}_{j} & d^i_j \; \text{support gold answer} , \\ 
q^{i,-}_{j} & \text{else},
\end{cases}
\end{equation}
where $d_{j}^{i}$ represents the documents retrieved by the query $q_{j}^{i}$.

\subsubsection{Planning Learning}\label{sec:plan_learn}

Based on the judgements, we employ $\mathcal{M}_{\theta}$ to perform supervised fine-tuning (SFT) first, followed by direct preference optimisation (DPO) to further align with the multi-aspect knowledge base.

\paragraph{Supervised Fine-tuning.}

 For the case where more than 3 positive queries exist in a single source, we randomly select 3 of them. For the case where no positive queries exist for a single source, we leave this source empty, meaning it is not used. For the case where no positive queries exist in all sources, we filter the sample out, meaning that none of the sources can provide valuable information. To assess knowledge from all sources, up to 3 positive queries $q^+$ on each source are ensembled to construct the positive plan $P^+$ for each question. We perform standard SFT, whose training objective is formulated as:
 \begin{equation}
     \mathcal{L}_{\text{SFT}} = - \mathbb{E}_{\left(x, P^+\right) \sim \mathcal{D}^+} \log \mathcal{M}_{\theta} \left(P^+ \mid x\right)
 \end{equation}

\paragraph{Direct Perference Optimisation}

The planner $\mathcal{M}_\theta$, which builds upon the SFT, is further aligned to multiple source knowledge sources. To achieve this, we collect negative plans for each question in the same manner as the positive plans. The positive and negative plans are then paired for each question to construct the dataset $\mathcal{D}^{\pm}$. Importantly, we only retain samples where both positive and negative plans are available to ensure balanced training. Subsequently, we employ DPO learning~\cite{Rafailov2023}, a method shown to provide stable and effective alignment. The training objective is expressed as:
\begin{multline}
    \mathcal{L}_{\text{DPO}}= - \mathbb{E}_{(x, P^+, P^-) \sim \mathcal{D}^{\pm}} \log \sigma (r_\theta(x, P^+) \\ - r_\theta(x, P^-)),
\end{multline}
where 
\begin{equation}
r_\theta(x, P)=\beta \log \frac{\mathcal{M_{\theta}}(P \mid x)}{\mathcal{M}_{\text{SFT}}(P \mid x)}.
\end{equation}
We expect that this method will enable the planner to achieve precise alignment with complex knowledge sources, thereby facilitating effective multi-source planning.

%% file: 05_experiments.tex
\section{Experiments}

\subsection{Experimental Settings}

\subsubsection{Datasets}

\begin{table}[htbp]
    \centering
        \resizebox{\linewidth}{!}{
        \begin{tabular}{lcccc}
        \toprule[1pt]
            \bf Dataset & \bf Format & \bf \# Train & \bf \# Dev & \bf \# Test  \\
            \midrule
            MedQA & \multirow{8}{*}{\makecell[c]{Closed-ended \\ QA}} & 6000 & 1200 & 1273 \\
            MedMCQA & & 6000 & 1200 & 1200 \\
            MMLU-Med & & - & - & 1089 \\
            PubMedQA & & 417 & 83 & 500 \\
            BioASQ & & 584 & 116 & 782 \\
            SEER & & 2113 & 422 & 1000 \\
            DDXPlus & & 5000 & 1000 & 1000 \\
            MIMIC-IV-ED & & 2917 & 583 & 1000 \\
            \midrule
            LiveQA & \multirow{3}{*}{\makecell[c]{Open-ended \\ Generation}} & 343 & 68 & 104 \\
            MedicationQA & & 450 & 90 & 150 \\
            ExpertQA-Biomed & & 375 & 75 & 150 \\
            \midrule
            \bf Total &  & 24199 & 4837 & 8248 \\
        \bottomrule[1pt]
        \end{tabular}
        }
        \caption{The statistics of datasets in our experiments. The training and development sets are combined to expect universal planning capabilities.}
    \label{tab:dataset_stat}
\end{table}

\begin{table*}[htbp]
\renewcommand{\arraystretch}{0.9}
\centering
\resizebox{\textwidth}{!}{
\begin{tabular}{l l c c c c c c c c >{\columncolor{mylightred}}c}
\toprule[1pt]
 & & \multicolumn{3}{c}{\textbf{Reasoning QA}} & \multicolumn{2}{c}{\textbf{Research QA}} & \multicolumn{3}{c}{\textbf{Clinical QA}} & \cellcolor{white} \\
 \cmidrule(lr){3-5}\cmidrule(lr){6-7}\cmidrule(lr){8-10}
\multirow{-2}{*}{\textbf{Planner}}  & \multirow{-2}{*}{\textbf{Method}}  & \bf MedQA & \bf MedMCQA & \bf MMLU. & \bf PubMedQA & \bf BioASQ & \bf SEER & \bf DDXPlus & \bf MIMIC-IV. & \bf \cellcolor{white} \multirow{-2}{*}{\makecell[c]{Average~}} \\
\midrule

\multicolumn{11}{c}{\textit{Reader: Frozen Qwen2.5-7B}} \\
\midrule
\multirow{3}{*}{-} & No Retrieval & 60.80 & 56.17 & 76.95 & 34.60 & 74.81 & 51.00 & 42.80 & 58.50 & 56.95 
\\
& Original Question & 62.45 & 63.25 & 80.90 & 47.00 & 89.00 & 58.40 & 42.80 & 57.90 & 62.71 
\\
& Query2Doc & 62.92 & 66.42 & 80.26 & 46.40 & 88.24 & \underline{58.80} & 42.40 & 56.90 & 62.79 
\\
\midrule
\multirow{3}{*}{\makecell[l]{Frozen \\ Qwen2.5-72B}} & Prompting & 72.11 & 65.33 & 81.73 & 53.80 & 89.64 & 57.10 & 48.70 & 62.00 & 66.30 
\\
& Reflexion & \underline{73.13} & 66.00 & 79.06 & 52.60 & 89.64 & 57.90 & 49.40 & 62.60 & 66.29 
\\
& SeRTS & 70.70 & \underline{66.83} & \underline{82.55} & \underline{55.60} & \textbf{90.03} & 57.10 & \underline{51.20} & 62.50 & \underline{67.06} 
\\
\midrule
\multirow{3}{*}{\makecell[l]{Trained \\ Qwen2.5-7B}} & Trainable Planning & 72.03 & 66.42 & 82.19 & 54.80 & \underline{89.90} & 57.20 & 46.40 & 60.30 & 66.16 
\\
& RaFe Planning & 70.86 & 66.50 & 78.70 & 53.40 & 89.77 & 55.20 & 50.30 & \underline{63.70} & 66.05 
\\
& \bf SPO Planning & \textbf{76.98} & \textbf{71.08} & \textbf{85.49} & \textbf{60.20} & 89.77 & \textbf{61.90} & \textbf{52.40} & \textbf{69.60} & \textbf{70.93} 
\\
\midrule

\multicolumn{11}{c}{\textit{Reader: Frozen Llama3.1-8B}} \\
\midrule
\multirow{3}{*}{-} & No Retrieval & 65.99 & 59.50 & 76.58 & 56.20 & 81.97 & 57.00 & 38.80 & 58.60 & 61.83 
 \\
& Original Question & 60.57 & 57.50 & 72.18 & 74.20 & 87.47 & 57.60 & 39.00 & 58.10 & 63.33 
 \\
& Query2Doc & 61.04 & 59.92 & 72.91 & \underline{74.80} & 87.21 & 57.60 & 39.20 & 57.70 & 63.80 
\\
\midrule
\multirow{3}{*}{\makecell[l]{Frozen \\ Qwen2.5-72B}} & Prompting & 71.17 & 62.08 & 75.94 & 71.40 & 89.00 & 57.50 & 41.10 & 58.60 & 65.85 
\\
& Reflexion & \underline{72.82} & 63.17 & 75.48 & 71.80 & \underline{89.51} & 56.80 & 41.40 & 59.70 & 66.34 
\\
& SeRTS & 71.88 & 63.25 & \underline{77.13} & 71.60 & \underline{89.51} & 57.00 & 42.90 & 60.10 & \underline{66.67}
\\
\midrule
\multirow{3}{*}{\makecell[l]{Trained \\ Qwen2.5-7B}} & Trainable Planning &  71.64 & 62.33 & 75.76 & 72.20 & 89.00 & 58.30 & \underline{43.60} & \underline{60.20} & 66.63 
\\
& RaFe Planning & 69.76 & \underline{63.67} & 74.75 & 72.20 & 89.00 & \underline{58.70} & 42.40 & 59.40 & 66.24 
\\
& \bf SPO Planning & \textbf{77.45} & \textbf{69.25} & \textbf{78.97} & \textbf{75.60} & \textbf{89.64} & \textbf{60.70} & \textbf{45.70} & \textbf{64.10} & \textbf{70.18} 
\\
\midrule

\multicolumn{11}{c}{\textit{Reader: Frozen Mistral0.3-7B}} \\
\midrule
\multirow{3}{*}{-} & No Retrieval & 49.18 & 45.58 & 63.54 & 47.40 & 73.27 & 49.40 & 23.50 & 58.20 & 51.26 
 \\
& Original Question & 53.18 & 58.00 & 68.96 & \textbf{67.00} & \textbf{88.87} & 51.10 & 30.60 & 57.00 & 59.34 
 \\
& Query2Doc & 52.40 & \underline{59.67} & 70.71 & \underline{66.60} & 86.83 & 52.90 & 30.50 & 57.00 & 59.58 
\\
\midrule
\multirow{3}{*}{\makecell[l]{Frozen \\ Qwen2.5-72B}} & Prompting & 64.26 & 57.75 & 72.73 & 59.80 & \underline{88.75} & 55.10 & 41.10 & 58.60 & 62.26 
\\
& Reflexion & 64.26 & 57.50 & 71.44 & 58.40 & 87.21 & \underline{56.50} & 41.70 & 58.50 & 61.94 
\\
& SeRTS & \underline{65.59} & 58.83 & \underline{73.00} & 58.60 & 87.34 & 56.40 & \underline{43.50} & 58.90 & \underline{62.77} 
\\
\midrule
\multirow{3}{*}{\makecell[l]{Trained \\ Qwen2.5-7B}} & Trainable Planning & 62.45 & 55.75 & 71.63 & 56.80 & 87.08 & 53.30 & 41.70 & 58.40 & 60.89 
\\
& RaFe Planning & 61.67 & 55.25 & 71.99 & 56.20 & 82.35 & 56.20 & 41.20 & \underline{60.40} & 60.66 
\\
& \bf SPO Planning & \textbf{71.09} & \textbf{64.25} & \textbf{75.02} & 65.20 & 87.60 & \textbf{59.00} & \textbf{45.20} & \textbf{64.80} & \textbf{66.52} 
\\
\bottomrule[1pt]
\end{tabular}
}
\vspace{-0.1cm}
\caption{Comparison of our method with baselines. \textbf{Bold} represents the best result, while \underline{underlining} represents the second-best result. All models are in the Instruct version. AWQ is applied to the models of 72B size. } 
\label{tab:main_result}
\end{table*}

\begin{table}[htbp]
\centering
    \resizebox{\linewidth}{!}{
    \begin{tabular}{lcccc} 
        \toprule[1pt]
        \bf Method & \bf Relevance & \bf Completeness & \bf Proficiency & \bf Interpretation \\
\midrule
\multirow{2}{*}{No Retrieval} & \bf 3.83  & 3.41  & 3.02  & -
\\
 & \small (4.1/3.7/3.6) & \small (3.3/3.6/3.4) & \small (2.8/3.6/2.7) & - 
\\
\midrule
\multirow{2}{*}{\makecell[l]{Original \\ Question}} & 3.79  & 3.55  & 3.48  & 3.29 
\\
& \small (4.1/3.5/3.7) & \small (3.6/3.5/3.6) & \small (3.5/3.3/3.6) & \small (3.0/3.2/3.7)
\\
\midrule
\multirow{2}{*}{SeRTS} & 3.80  & 3.49  & 3.30  & 3.19 
\\
& \small (3.9/3.9/3.6) & \small (3.3/3.8/3.3) & \small (3.6/3.6/2.7) & \small (3.4/3.0/3.1)
\\
\midrule
\bf \multirow{2}{*}{\makecell[l]{SPO \\ Planning}} & 3.79  & \bf 4.18  & \bf 4.08  & \bf 3.97 
\\
& \small (4.1/3.7/3.6) & \small (4.3/4.0/4.3) & \small (4.2/4.0/4.1) & \small (3.8/4.1/4.0)
\\

        \bottomrule[1pt]
    \end{tabular}
    }
    \caption{Expert evaluation results of open-ended generation datasets. The reader is frozen Qwen2.5-7B. Each cell reports the average value (LiveQA/MedicationQA/ExpertQA-Biomed).}
    \label{tab:expert}
\end{table}

We conduct our experiments on 11 datasets, which are well-defined and widely adopted for testing LLMs, as shown in Table~\ref{tab:dataset_stat}. These tasks represent the core focus of the medical application (detailed in Table~\ref{tab:dataset_app}). Reasoning QA datasets include MedQA~\cite{Jin2020}, MedMCQA~\cite{Pal2022}, and MMLU-Med~\cite{Hendrycks2021}. Research QA datasets include PubMedQA~\cite{Jin2019} and BioASQ~\cite{Tsatsaronis2015}. Clinical QA datasets include SEER~\cite{Dubey2023}, DDXPlus~\cite{Tchango2022} and MIMIC-IV-ED~\cite{Xie2022}. Long-form Answering datasets include LiveQA~\cite{Abacha2017}, MedicationQA~\cite{Abacha2019} and ExpertQA-Biomed~\cite{Malaviya2024}. More details of the datasets can be found in Appendix~\ref{appendix:train_test_data}. The accuracy is adopted as the metric for closed-ended QA datasets, while expert evaluation is used for open-ended generation datasets. The details of expert evaluation are shown in Appendix~\ref{appendix:expert}.

\subsubsection{Retrieval Procedure}
The proposed MedOmniKB is adopted as our retrieval knowledge base. For the unstructured sources, the format of the queries is \texttt{``query0; query1; ...''}. We retrieve the top-20 documents and then rerank the top-10 documents using MedCPT models~\cite{Jin2023}. For the structured source, the format of the queries is \texttt{``term0, query0; term0, query1; ...''}. We search for the definition of each \texttt{``term''} and all its reranked top-10 relations. The latter half of the search results is excluded during the planning judging phase to enhance efficiency.

\subsection{Baselines}

\textbf{Given the limited research on efficient multi-source planning for medical information retrieval, we replicate single-source querying approaches as baselines.} Unsupervised methods include No Retrieval, Original Question~\cite{Xiong2024}, Query2Doc~\cite{Wang2023}, Prompting~\cite{Ma2023}, Reflexion~\cite{Shinn2023}, and SeRTS~\cite{Hu2024}, all of which are adapted to support multi-source planning. Supervised methods consist of Trainable Planning~\cite{Ma2023} and RaFe Planning~\cite{Mao2024}, implemented with modifications by adjusting the judgment signal, specifically using QA accuracy (Rouge-L~\cite{Lin2004} for open-ended generation) and rerank scores for query evaluation, respectively. Additional details on the baseline methods can be found in Appendix~\ref{appendix:baselines}.

\begin{table}[t]
\centering
    \resizebox{\linewidth}{!}{
    \begin{tabular}{llll} 
        \toprule[1pt]
        \bf Sources & \bf MedQA & \bf PubMedQA & \bf SEER \\
        \midrule
All & 76.98 & 60.20 & 61.90 \\
~- Book & 70.38 \textcolor{purple}{(-8.57\%)} & 52.60 \textcolor{purple}{(-12.62\%)} & 56.80 \textcolor{purple}{(-8.24\%)}
\\
~- Guideline & 72.35 \textcolor{purple}{(-6.01\%)} & 56.40 \textcolor{purple}{(-6.31\%)} & 52.10 \textcolor{purple}{(-15.83\%)}
\\
~- Research & 71.72 \textcolor{purple}{(-6.83\%)} & 35.40 \textcolor{purple}{(-41.20\%)} & 51.20 \textcolor{purple}{(-17.29\%)}
\\
~- Wiki & 72.35 \textcolor{purple}{(-6.01\%)} & 58.20 \textcolor{purple}{(-3.32\%)} & 55.50 \textcolor{purple}{(-10.34\%)}
\\
~- Graph & 71.48 \textcolor{purple}{(-7.14\%)} & 58.20 \textcolor{purple}{(-3.32\%)} & 56.80 \textcolor{purple}{(-8.24\%)}
\\
\midrule
No & 60.80 \textcolor{purple}{(-21.02\%)} & 34.60 \textcolor{purple}{(-42.52\%)} & 51.00 \textcolor{purple}{(-17.61\%)}
\\
        \bottomrule[1pt]
    \end{tabular}
    }
    \caption{Accuracy and \textcolor{purple}{relative drop ratio} of reader when dropping documents from each source.}
    \label{tab:drop_src}
\end{table}

\subsection{Main Results}

To evaluate the effectiveness of our method, we incorporate three frozen readers with varying intrinsic knowledge: Qwen2.5-7B~\cite{QwenTeam2024}, Llama3.1-8B~\cite{Dubey2024}, and Mistral0.3-7B~\cite{Jiang2023}. The use of these pretrained models is consistent with their intended purposes. The main results are summarized in Table~\ref{tab:main_result} and Table~\ref{tab:expert}, leading to the following key observations:

\noindent\textbf{1. Effectiveness of Planning Methods:} Planning approaches (baselines excluding Original Question and Query2Doc) consistently outperform direct retrieval methods that rely on the original question or pseudo-document queries across most test cases. This underscores the importance of multi-source planning, which enables language models to retrieve information dynamically from diverse sources. However, for PubMedQA and BioASQ, non-planning approaches yield competitive results. This can be attributed to the strong alignment between these datasets and the ``Research'' source, reducing the need for planning.
    
\noindent\textbf{2. Superiority of SPO Training Signals:} The SPO method demonstrates superior training signals compared to existing trainable approaches. SPO avoids the risk of intrinsic knowledge overshadowing retrieved knowledge~\cite{Ma2023}. Additionally, it provides more accurate judgments than those proposed in \citet{Mao2024}, further validating the effectiveness of our method.
    
\noindent\textbf{3. Performance Across Readers:} When paired with each reader, SPO planning achieves the best performance in most configurations, outperforming models with up to 10 times the number of parameters. Furthermore, SPO can maximally enhance the completeness, proficiency, and interpretation of LLM medical responses. These improvements highlight SPO planning’s ability to provide readers with abundant and relevant knowledge.

\subsection{Enhancement for Different Sizes of Readers}

\begin{figure}[t]
    \centering
    \includegraphics[width=0.9\linewidth]{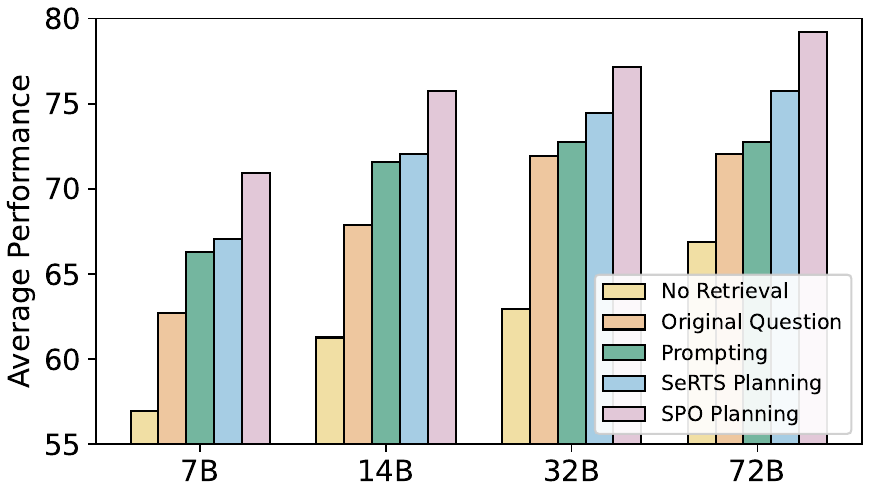}
    \caption{Average accuracy of different sizes of readers paired with different planners.}
    \label{fig:diff_readers}
\end{figure}

We also use models of different sizes as readers. We use frozen 7B, 14B, 32B, and 72B Qwen2.5-Instruct models as readers and show the average accuracy on all closed-ended datasets as shown in Figure~\ref{fig:diff_readers}. AWQ is applied to models of 32B and 72B sizes. In general, the performances of the readers in all cases rise with increasing model size, which is attributed to the increasing intrinsic knowledge. Significantly, it can be found that the SPO planner brings the most enhancements among all planners, even as the readers' intrinsic knowledge increases.

\subsection{Effectiveness of Different Sources}

To investigate the effects of each retrieval source, we systematically exclude documents from each source which is retrieved using the SPO planner. The results are shown in Table~\ref{tab:drop_src}. We utilise the frozen Qwen2.5-7B-Instruct model as the reader. Overall, every retrieval source enhances the ability to answer medical questions to varying degrees, with the “Book”, “Guideline”, and “Research” sources providing the most significant improvements.
Regarding the impact of sources on different types of QA, the “Research” source notably impacts performance on the PubMedQA dataset. It aligns with the “Research” style inherent to the dataset. The “Guideline” and “Research” sources have a significant impact on SEER, which can be attributed to its characteristics as a treatment planning dataset.
Moreover, the visualization of source planning statistics is shown in Appendix~\ref{appendix:vis_sp}.

\subsection{Effectiveness of Training Stages}

To examine the impact of the two training phases, SFT and DPO, we present the results of the planner model trained under different configurations in Table~\ref{tab:drop_sft_dpo}. Frozen Qwen2.5-7B-Instruct is used as the reader. It is important to note that the Frozen model referenced here is 7B, whereas the Frozen model in Table~\ref{tab:main_result} is 72B. The results demonstrate that SFT significantly enhances the model’s multi-source planning capability, providing a strong foundational ability. The subsequent DPO phase further refines and strengthens this capability. However, applying DPO directly to the planner yields only limited improvements, underscoring the critical role of the SFT phase in establishing the model’s initial planning proficiency.

\begin{figure}[t]
    \centering
    \includegraphics[width=1\linewidth]{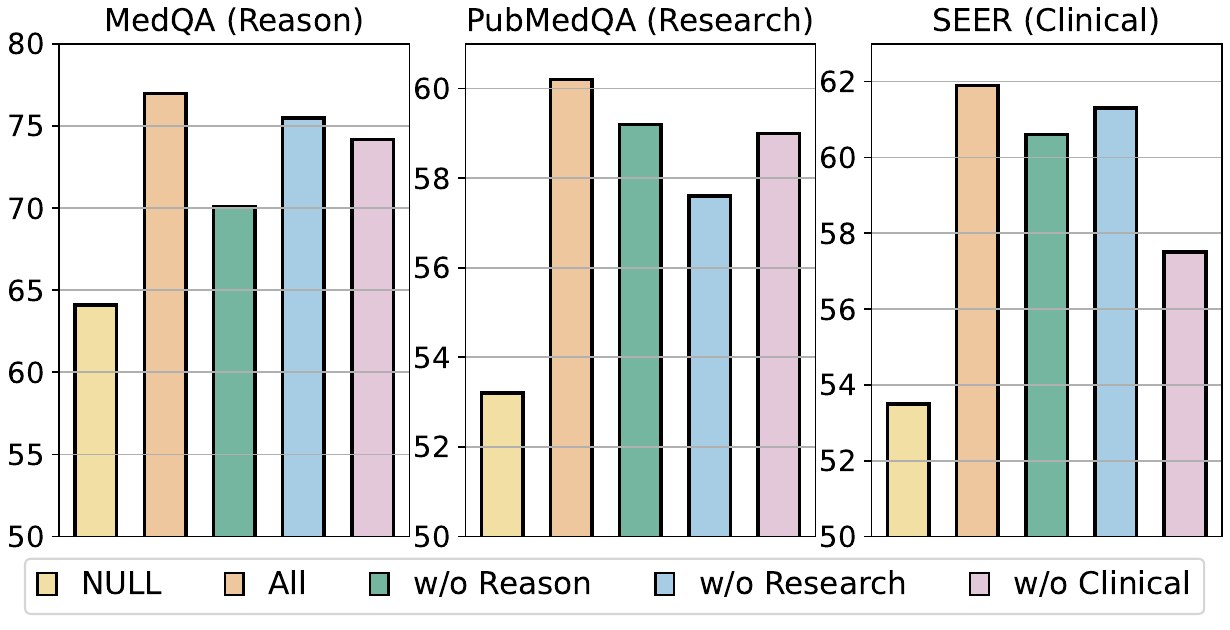}
    \caption{Accuracy of the reader paired with SPO planner trained using different categories of data.}
    \label{fig:robust_type}
\end{figure}

\begin{table}[t]
    \centering
        \resizebox{\linewidth}{!}{
        \begin{tabular}{lccc}
        \toprule[1pt]
            \bf Training Strategy & \bf MedQA & \bf PubMedQA & \bf SEER \\
            \midrule
Frozen & 64.10  & 53.20  & 53.50 
 \\
 ~+ SFT & 74.08  & 59.20  & 58.50
 \\
 ~+ SFT + DPO & \bf 76.98  & \bf 60.20  & \bf 61.90
 \\
 ~+ DPO & 67.48  & 55.80  & 54.30 
 \\
        \bottomrule[1pt]
        \end{tabular}
        }
        \caption{Accuracy of the reader paired with SPO planner trained using different strategies.}
    \label{tab:drop_sft_dpo}
\end{table}

\begin{figure}[t]
    \centering
    \includegraphics[width=0.8\linewidth]{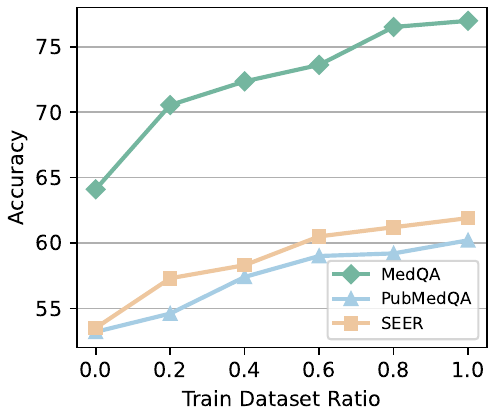}
    \caption{Accuracy of the reader paired with SPO planner trained using different ratios of data.}
    \label{fig:robust_ratio}
    \vspace{-0.2cm}
\end{figure}

\begin{table}[t]
    \centering
        \resizebox{\linewidth}{!}{
        \begin{tabular}{lccc}
        \toprule[1pt]
            \bf Training Strategy & \bf MedQA* & \bf PubMedQA* & \bf SEER* \\
            \midrule
No Retrieval & 60.80 & 34.60 & 51.00 \\
Frozen & 62.77 & 47.00 & 52.50 \\
~+ SPO & 69.21 & 58.20 & 56.80 \\
~+ SPO (OOD corpus) & 65.44 & 50.60 & 53.20 \\
        \bottomrule[1pt]
        \end{tabular}
        }
        \caption{Accuracy of the reader paired with SPO planner trained using different strategies. The asterisk (*) indicates that only a single retrieval source is available during the testing phase. Specifically, only the Book source is available on MedQA, only the Research source is available on PubMedQA, and only the Guideline source is available on SEER.}
    \label{tab:ood_retrieval_source}
\end{table}

\subsection{Adaptability to Out-of-Distribution Data}

We classify the out-of-distribution (OOD) task into three categories according to Table~\ref{tab:dataset_app}, where we systematically exclude the training data for individual categories and assess the planner’s OOD performance. We use the frozen Qwen2.5-7B-Instruct as the reader. Figure~\ref{fig:robust_type} compares the performance of models without training, trained with the full dataset, and trained with data from each type of data excluded. 
Firstly, the SPO planner outperforms the untrained planner across the test sets of all three domains under the out-of-type setting (for example, MedQA in the "w/o Reason" case). This demonstrates that the SPO method can learn a generalized source planning ability, regardless of the category of training data. Furthermore, it can be observed that when training data of a different type from the current test set is dropped (for example, MedQA in the "w/o Research" case), the model's performance slightly declines. This further confirms that all types of data contribute to learning a universal planning ability.

\subsection{Adaptability to Out-of-Distribution Retrieval Source}

To explore the SPO's adaptability to an unknown retrieval corpus, we designed a new baseline, SPO (OOD corpus). We select “Book” as the unknown corpus for MedQA, “Research” for PubMedQA, and “Guideline” for SEER. During training, only the unknown source is excluded in the prompt and label (a total of 4 sources); however, during testing, only this source is included in the prompt (a total of 1 source). All methods' test prompts remain the same. Frozen Qwen2.5-7B is adopted as the reader.

Table~\ref{tab:ood_retrieval_source} shows that the SPO (OOD corpus) planner can retrieve more closely aligned with the characteristics of unknown sources compared to the frozen planner. The SPO method enhances the perception of the source, even though the source is not included in the training set. It is reasonable, as all sources share semantic similarities in the medical field. The adaptability of the SPO method to OOD corpora broadens its potential applications.

\subsection{Robustness to Training Data Quantity}\label{sec:robust_quantity}

We further examine the impact of data quantity by sampling different proportions of the training dataset and applying the same experimental setup as in the main study. We use the frozen Qwen2.5-7B-Instruct as the reader. As shown in Figure~\ref{fig:robust_ratio}, planning performance improves consistently with increasing data volume but begins to plateau at higher levels. These results indicate that the SPO method significantly enhances planning capability even with limited data, while larger datasets provide diminishing yet notable additional benefits.

%% file: 06_conclusion.tex
\section{Conclusion}

This study tackles the challenge of multi-source planning in healthcare by introducing the MedOmniKB knowledge base and proposing the Source Planning Optimisation method. The MedOmniKB combines diverse medical information sources with high relevance. The Source Planning Optimisation method significantly enhances language models' ability to retrieve from various knowledge sources. Extensive experiments show our optimised small model has outperformed the larger one in planning efficiency. In this work, we underscore the importance of effective source planning. Our model effectively integrates multi-source knowledge, advancing the application of language models in the healthcare field.

%% file: 07_limit_ethi.tex
\section*{Limitations}

There are some limitations in our work. First, our constructed MedOmniKB may not encompass all medical knowledge resources. However, the five sources we select are not only representative but also essential for various applications requiring medical knowledge. Other medical knowledge not covered by the five sources still needs to be identified and incorporated into MedOmniKB.

Moreover, the exploration and judging steps in our SPO approach incur a relatively large retrieval and inference cost. The retrieval cost arises from the large and diverse retrieval repository, while the inference cost is due to the separate evaluation for each document set. In our opinion, the accurate judgement of each query is essential for the model to learn source planning. We alleviate the cost problems by optimising the retrieval framework based on Qdrant and multi-process parallelism. Comparison of cost with existing supervised methods and cost analysis in multi-process scenarios are presented in Appendix~\ref{appendix:cost}.

Lastly, there are limitations in our evaluation datasets and methods. Although we have included as many categories of medical datasets as possible, there may still be unexplored specific medical scenarios. Additionally, while we have employed evaluation methods such as accuracy for question-answering, and expert assessments for long-text generation, further evaluation is still needed. This should include assessing user satisfaction, treatment outcomes, and other factors in real medical scenarios.

\section*{Ethical Consideration}

The deployment of large language models (LLMs) in healthcare raises significant ethical challenges that warrant careful consideration. Ensuring accuracy is paramount, as inaccuracies may adversely affect patient outcomes. Our Source Planning Optimisation (SPO) method addresses this concern by enhancing the retrieval of information from credible sources. It can help models generate accurate and reliable responses.

Furthermore, transparency in model outputs is markedly improved through access to the retrieved knowledge. By documenting the sources of information utilized in generating responses, healthcare professionals can understand the foundations of the information presented to them. This transparency also enables practitioners to assess the reliability of the model responses.

%% file: 08_appendix.tex
\section{Additional Experiments}
\subsection{Optimal Number of Queries in Planning Exploration}~\label{appendix:diversity}

\begin{figure}[htbp]
    \centering
    \includegraphics[width=\linewidth]{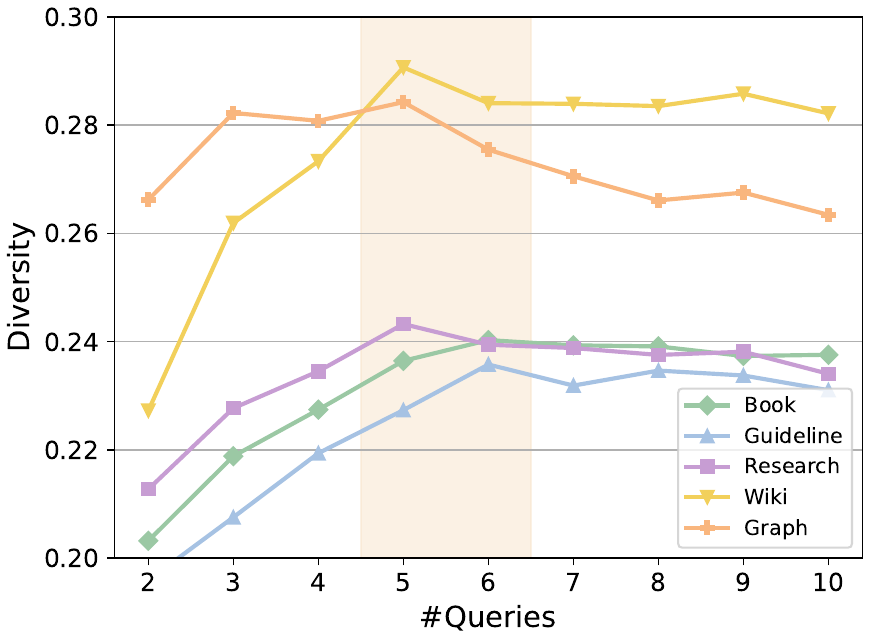}
    \caption{The query diversity of varying queries on the development set.}
    \label{fig:query_div}
\end{figure}

\begin{figure*}[htbp]
    \centering
    \includegraphics[width=0.9\linewidth]{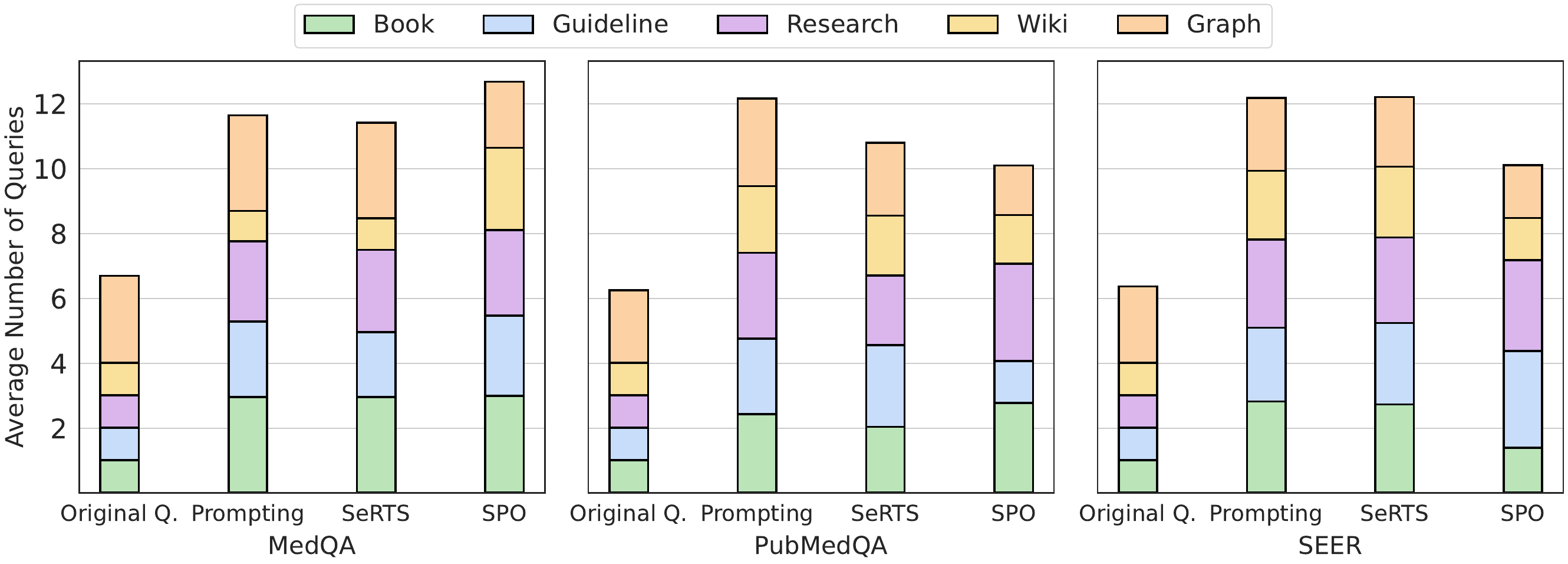}
    \caption{Average number of queries of the SPO method for different sources on test sets. ``Original Q.'' represents ``Original Question''. Note that the maximum number of queries per source is 3, catering for the LLM context length limit.}
    \label{fig:num_queries}
\end{figure*}

\begin{table*}[htbp]
    \centering
\resizebox{0.9\linewidth}{!}{
    \begin{tabular}{llccccc} 
        \toprule[1pt]
        \bf Method & \bf LLM Judger & \bf Accuracy & \bf Precision & \bf Recall & \bf F1 \\
        \midrule
Trainable Planning & - & 64.50 & 43.29 & 82.74 & 56.84 \\
RaFe Planning & - & 63.88 & 64.12 & 67.40 & 65.72 \\
\midrule
SPO Planning & Qwen2.5-32B & 89.75 & 92.13 & 89.24 & 90.66 \\
& Llama3.1-72B & 90.88 & 90.74 & 92.24 & 91.48 \\
& DeepSeek-R1-Distill-Llama-70B & 93.50 & 93.06 & 94.81 & 93.93 \\
& Qwen2.5-72B & 93.00 & 91.67 & 95.19 & 93.40 \\
        \bottomrule[1pt]
    \end{tabular}
}
    \caption{Human evaluation of existing methods for constructing query training data. All models are in the Instruct version and quantized using AWQ.}
    \label{tab:human_eval_judge}
\end{table*}

In this section, we aim to identify the optimal number of queries for each source in Planning Exploration~(\ref{sec:plan_exp}). Our objective is to balance the need for diverse queries with the constraints of retrieval and inference costs, which can become significant with excessive querying. To achieve this, we define the optimal number based on the diversity of the source’s queries, as our primary goal is to ensure diverse exploration. Specifically, the diversity of a source’s queries ${ q^i }$ is quantified as:
\begin{equation}
    D = \frac{1}{\binom{N}{2}} \sum_{j=1}^{N} \sum_{k=j+1}^{N} \operatorname{distance}(q^i_j, q^i_k),
\end{equation}
where $N$ is the number of $\{q^i\}$ and $\operatorname{distance} (\cdot, \cdot)$ is the cosine distance between encoded query vectors.

Figure~\ref{fig:query_div} illustrates the query diversity across different sources on the development set. In the SPO approach, we set the number of queries per source to 6, as the diversity peaks when the number of queries ranges between 5 and 6.

\subsection{Human Evaluation of LLM Judgements} \label{sec:human_judge}

We present a manual evaluation of existing methods for constructing query training data. We randomly sample 800 samples from the training set. Medical researchers are employed to determine whether each document supports the correct answer, yielding 800 tuples of (question + correct answer, document, human judgement). Subsequently, we treat the discrimination results of different methods as hypotheses, and human judgment as references and calculate accuracy, precision, recall and F1 scores.

The human evaluation results are illustrated in Table~\ref{tab:human_eval_judge}. It is clear that the SPO method yields much superior query quality than the existing methods. As mentioned earlier, Trainable Planning~\cite{Ma2023} evaluates query value based on downstream task performance, potentially influenced by LLMs' intrinsic knowledge. RaFe Planning~\cite{Mao2024} uses rerank scores directly, offering a quick method but mainly focused on relevance ranking. Moreover, our findings indicate that LLMs possess a sufficient capacity to perform the judgment task, even with the smaller Qwen2.5-32B-Instruct model. This illustrates the superiority of the SPO method.

\begin{table*}[htbp]
    \centering
        \resizebox{\linewidth}{!}{
        \begin{tabular}{lcccccc}
        \toprule[1pt]
 & \multicolumn{2}{c}{\textbf{MedQA}} & \multicolumn{2}{c}{\textbf{PubMedQA}} & \multicolumn{2}{c}{\textbf{SEER}} \\
 \cmidrule(lr){2-3}\cmidrule(lr){4-5}\cmidrule(lr){6-7}
\multirow{-2}{*}{\textbf{Training Strategy}} & Uncertainty ($\downarrow$) & Accuracy ($\uparrow$) & Uncertainty ($\downarrow$) & Accuracy ($\uparrow$) & Uncertainty ($\downarrow$) & Accuracy ($\uparrow$) \\
\midrule
Frozen planner & 0.343 & 64.10 & 0.214 & 53.20 & 0.434 & 53.50 \\
~+ SFT ($P^{+}$) & 0.325 & 74.08 & 0.201 & 59.20 & 0.432 & 58.50 \\
~+ SFT ($P^{+}$) + DPO ($P^{\pm}$) & \bf 0.276 & \bf 76.98 & \bf 0.185 & \bf 60.20 & \bf 0.386 & \bf 61.90 \\
~+ SFT ($P^{-}$)& 0.563 & 61.90 & 0.339 & 46.40 & 0.599 & 52.40 \\
        \bottomrule[1pt]
        \end{tabular}
        }
        \caption{Uncertainty and accuracy of the reader paired with planners trained using different strategies.}
    \label{tab:uncertainty}
\end{table*}

\begin{table*}[htbp] \small
    \centering
        \begin{tabular}{lcccc}
        \toprule[1pt]
            \bf Cases & \bf \bf \makecell[c]{\# GPUs \\ (\# Processes)} & \bf \makecell[c]{Retrieval Time \\ per Sample (s)} & \bf \makecell[c]{Inference Time \\ per Sample (s)} & \bf \makecell[c]{Estimated Total Time on \\ Training/Test Dataset (h)} \\
            \midrule
            \multirow{5}{*}{\makecell[l]{Training Data \\ Construction \\ (24199 samples)}} & 1 & 7.1 & 60.1 & 451.7 \\
            & 2 & 4.2 & 30.1 & 230.6 \\
            & 4 & 3.0 & 15.8 & 126.4 \\
            & 8 & 2.8 & 8.3 & 74.6 \\
            & 16 & 2.7 & 4.4 & 47.7 \\
            \midrule
            \multirow{4}{*}{\makecell[l]{Testing \\ Using SPO Model \\ (8248 samples)}} & 1 & 4.1 & 2.1 & 14.2 \\
            & 2 & 2.2 & 1.1 & 7.6 \\
            & 4 & 1.4 & 0.6 & 4.6 \\
            & 8 & 1.2 & 0.3 & 3.4 \\
        \bottomrule[1pt]
        \end{tabular}
        \caption{The detailed cost analysis for the training data construction and the test process. During training, inference involves the Qwen2.5-72B-Instruct-AWQ model generating multiple queries and judging their corresponding document sets. In testing, inference involves the trained Qwen2.5-7B-Instruct model generating multiple queries. Note that the time taken by the Reader to respond during testing is not included in this table, as it varies with the Reader's size and prompts.}
    \label{tab:cost_multi}
\end{table*}

\subsection{Assessment of Uncertainty Introduced by Retrieval}
Experiments are conducted to assess the uncertainty brought by the SPO method. The uncertainty of Reader responses reflects the ambiguity present in the retrieved information. We reimplement Long-Text Uncertainty Quantification (LUQ)~\cite{Zhang2024} to assess the uncertainty of LLMs. The temperature is set to 0.7, and the number of samples is set to 9. As for the “accuracy”, we set the generation temperature to 0. Frozen Qwen2.5-7B is adopted as the Reader. We report the QA uncertainty and accuracy in different training strategies.

As illustrated in Table~\ref{tab:uncertainty}, we can find that both phases of SPO training (SFT and DPO) reduce the uncertainty of retrieved content and improve QA accuracy. However, the uncertainty becomes higher, and QA accuracy becomes lower when the planner is trained only on negative plans. It indicates that the LLM judger has considered the ambiguous and uncertain information in the retrieval results.

\subsection{Human Evaluation of Long-form Generation} \label{appendix:expert}

Medical researchers (graduate students) in Shanghai are instructed to rate the predictions for long-form generation questions (LiveQA, MedicationQA, and ExpertQA-Biomed) using a reference solution. The annotators are informed that the annotation is conducted for academic purposes only, and they participate in the task voluntarily without special payments required. Following \citet{Wang2024d, Liao2024a}, each response is scored across four aspects—Relevance, Completeness, Medical Proficiency, and Interpretability—on a grading scale from 1 to 5. The instructions are shown in Instruction~\ref{prompt:human_rate}.

\section{Additional Analysis}

\subsection{Visualization of Planning Statistics}\label{appendix:vis_sp}

Figure~\ref{fig:num_queries} visualises the average number of queries generated by the baselines (Original Question, Prompting, and SeRTS) and our SPO method. This comparison highlights the extent to which the planners prioritise different sources. In terms of total queries, SPO generates more queries on MedQA than the baselines, reflecting the dataset’s need for comprehensive knowledge as it deals with diverse topics of medical diagnosis. Conversely, for PubMedQA and SEER, which involve specialised medical knowledge, SPO generates more concise queries than the baselines. Additionally, regarding the planner’s focus on sources, SPO provides a more balanced query distribution for MedQA. It allocates more focused attention to the ``Research'' source in PubMedQA and the ``Guideline'' source in SEER, aligning with their corresponding real requirements.

\subsection{More Planning Statistics}
We count the number of words per query and compute the semantic diversity of queries per plan. The results are averaged across all test samples. The computation of semantic diversity follows the way in Appendix~\ref{appendix:diversity}.

\begin{table}[t]
    \centering
    \resizebox{\linewidth}{!}{
    \begin{tabular}{lccccccccc} 
        \toprule[1pt]
        \textbf{\# Words} & \bf Book & \bf Guideline & \bf Research & \bf Wiki & \bf Graph & \bf Average \\
        \midrule
Original & 117.1 & 117.1 & 117.1 & 117.1 & 119.5 & 117.6 \\
Prompting & 5.7 & 6.5 & 7.0 & 3.9 & 5.8 & 5.8 \\
SeRTS & 6.5 & 6.7 & 7.0 & 4.1 & 5.6 & 6.0 \\
SPO Planning & 5.4 & 6.8 & 6.9 & 3.9 & 6.6 & 6.0 \\
        \bottomrule[1pt]
    \end{tabular}
    }
    \caption{Statistics of the number of words per query.}
    \label{tab:stat_words_plan}
\end{table}

\begin{table}[t]
    \centering
    \resizebox{\linewidth}{!}{
    \begin{tabular}{lccccccccc} 
        \toprule[1pt]
        \textbf{Diversity} & \bf Book & \bf Guideline & \bf Research & \bf Wiki & \bf Graph & \bf Average \\
        \midrule
Original & - & - & - & - & 0.301 & 0.301 \\
Prompting & 0.248 & 0.204 & 0.275 & 0.232 & 0.297 & 0.251 \\
SeRTS & 0.237 & 0.193 & 0.192 & 0.224 & 0.285 & 0.226 \\
SPO Planning & 0.220 & 0.173 & 0.218 & 0.207 & 0.287 & 0.221 \\
        \bottomrule[1pt]
    \end{tabular}
    }
    \caption{Diversity of queries per plan.}
    \label{tab:stat_diversity_plan}
\end{table}

\begin{table}[t]
    \centering
        \resizebox{\linewidth}{!}{
        \begin{tabular}{lccc}
        \toprule[1pt]
            \bf Method & \bf Retrieval Time & \bf Inference Time \\
            \midrule
Trainable Planning & 7.1 & 64.5
 \\
RaFe Planning & 7.1 & 2.3
 \\
SPO Planning & 7.1 & 60.1
 \\
        \bottomrule[1pt]
        \end{tabular}
        }
        \caption{Comparison of the cost of training data construction with existing supervised methods.}
    \label{tab:cost_compare}
\end{table}

As illustrated in Table~\ref{tab:stat_words_plan}, we can find that the number of words in ``Original Question'' is significantly higher compared to other planning methods, which hinders the acquisition of knowledge of a wide semantic range. Moreover, the SPO queries for the Wiki source have the fewest words, which is reasonable since the Wiki source contains encyclopaedic entries.

Table~\ref{tab:stat_diversity_plan} shows that the overall semantic diversity of SPO's queries is the lowest. This is meaningful since the SPO approach can obtain more focused information based on the source's characteristics compared to baselines. The queries constructed by baselines are more dispersed due to the inability to perceive the sources.

\subsection{Cost Analysis}\label{appendix:cost}

The environmental information is listed as follows:
\begin{itemize}
    \item \textbf{Retrieval.} No GPUs are required. (Qdrant supports CPU vector search acceleration)
    \item \textbf{Reranking.} Only 1 GPU is required (with a minimum of 2 GB of graphics memory needed for the efficiency of the reranking model).
    \item The above two modules are deployed using FastAPI, allowing MedOmniKB to handle multiple requests simultaneously.
    \item \textbf{Inference.} At least $1$ GPUs are required. We use the 80G NVIDIA A100 to run a single LLM under the vLLM framework. (The more GPUs available, the faster SPO can construct training data and perform inference across the entire dataset)
\end{itemize}

We randomly select 1000 training samples and 1000 test samples. First, we conduct the comparison of cost with existing supervised methods, as shown in Table~\ref{tab:cost_compare}. It can be observed that our SPO method remains competitive among supervised approaches, especially given that it offers the best supervision signals, as highlighted in Section~\ref{sec:human_judge}.

Then we measure the time spent on retrieval and inference in various multi-process cases. We divide the total time spent by the number of samples to get the average time spent. The multi-GPU (one process on each GPU) will reduce the average time spent. Finally, we multiply the average by the total number of samples to get the estimated time on the entire training/test dataset. The cost results are shown in Table~\ref{tab:cost_multi}. The process of constructing training data is computationally intensive, but is performed only once during model training. The actual cost during real-time inference is much lower, which is suitable for real-world, real-time use cases. For the training data construction process, we recommend 1) using 4-8 GPUs to execute the process, or alternatively, 2) reducing the number of training samples. In Section~\ref{sec:robust_quantity}, we demonstrate that the SPO method enhances planning capability even with limited training samples.

\begin{table}[t] \small
    \centering
    \begin{tabular}{cccccccccc} 
        \toprule[1pt]
        \textbf{Category} & \bf Count \\
        \midrule
Anatomy & 417 \\
Biochemistry & 1968 \\
Ethics and Law in Medicine & 254 \\
Immunology & 568 \\
Internal Medicine & 2268 \\
Microbiology & 755 \\
Neurology & 1713 \\
Obstetrics and Gynecology & 518 \\
Pathology & 1774 \\
Pediatrics & 648 \\
Pharmacology & 1194 \\
Physiology & 753 \\
Psychiatry & 1226 \\
Public Health & 1351 \\
Radiology & 1084 \\
Surgery & 1691 \\
\midrule
\bf Total & \bf 18182 \\
        \bottomrule[1pt]
    \end{tabular}
    \caption{Statistics of collected books.}
    \label{tab:stat_books}
\end{table}

\section{Case Studies}\label{appendix:case}

\begin{table*}[htbp]
\centering
\resizebox{\linewidth}{!}{
    \begin{tabular}{p{2.5cm}p{20.5cm}}
       \toprule[1pt]
        \multicolumn{2}{c}{\bf Case Study 1} \\
       \midrule
       \bf Original \newline Question & A 48-year-old man is brought to the emergency department for sudden onset of difficulty breathing 6 hours ago. For the past several months, he has had shortness of breath on exertion and while lying down on the bed, frequent headaches, and swelling of his feet. He does not take any medications despite being diagnosed with hypertension 10 years ago. His pulse is 90/min, respirations are 20/min, blood pressure is 150/110 mm Hg, and temperature is 37.0°C (98.6°F). Physical examination shows an overweight male in acute distress with audible wheezes. Crackles are heard bilaterally and are loudest at the lung bases. Which of the following findings on cardiac auscultation will most likely be present in this patient? \newline A. Loud P2. B. S3 gallop. C. Absent S4. D. A loud S1. \\
       \midrule
       \bf Prompting & 
       <book> Cardiac auscultation in heart failure ; Physical examination findings in heart failure ; S3 gallop in heart failure </book> \newline
       \textcolor{purple}{<guideline> Clinical presentation of heart failure ; Auscultation findings in heart failure</guideline>} \newline
       <research> Auscultation findings in patients with heart failure ; S3 gallop in heart failure </research> \newline
       \textcolor{purple}{<wiki> </wiki>} \newline
       <graph> heart failure , auscultation findings ; S3 gallop , clinical significance ; pulmonary edema , auscultation findings </graph>
       \\
       \midrule 
      \bf SeRTS & 
       <book> Cardiac auscultation in heart failure ; Physical examination findings in hypertensive heart disease ; S3 gallop in heart failure </book> \newline
       \textcolor{purple}{<guideline> Management of hypertensive heart disease ; Clinical presentation of heart failure ; Auscultation findings in patients with hypertension and heart failure </guideline>} \newline
       <research> Auscultation findings in patients with hypertensive heart failure ; S3 gallop as a marker of heart failure severity </research> \newline
       \textcolor{purple}{<wiki> </wiki>} \newline
       <graph> heart failure , auscultation findings ; hypertensive heart disease , cardiac auscultation </graph>
       \\
       \midrule
       \bf SPO Planning &
       <book> Cardiac auscultation in heart failure ; physical examination of heart sounds in hypertensive patients ; auscultation in patients with bilateral crackles </book> \newline
       \textcolor{mydrakgreen}{<guideline> clinical guidelines for heart failure management ; guidelines for diagnosing pulmonary edema ; cardiac auscultation in patients with heart failure </guideline>} \newline
       <research> Auscultation findings in patients with hypertensive heart failure ; auscultation in patients with shortness of breath </research> \newline
       \textcolor{mydrakgreen}{<wiki> cardiac auscultation ; heart failure symptoms ; pulmonary edema </wiki>} \newline
       <graph> heart failure , auscultation findings ; S3 gallop , clinical significance ; wheezing , auscultation findings </graph>
       \\
       \midrule
       \multicolumn{2}{c}{\bf Case Study 2} \\
       \midrule
       \bf Original \newline Question & Is the determination of specific IgE against components using ISAC 112 a reproducible technique? \newline A. yes. B. no. C. maybe. \\
       \midrule
       \bf Prompting & 
       <book> reproducibility of ISAC 112 for specific IgE determination ; validation studies of ISAC 112 ; reliability of ISAC 112 in clinical practice </book> \newline
       \textcolor{purple}{<guideline> reproducibility of ISAC 112 in clinical guidelines ; recommendations for ISAC 112 use in allergy testing ; guidelines on the reliability of component-resolved diagnostics </guideline>} \newline
       <research> reproducibility of ISAC 112 in peer-reviewed studies ; validation of ISAC 112 for specific IgE ; reliability and accuracy of ISAC 112 in allergy diagnosis </research> \newline
       <wiki> ISAC 112 reproducibility ; ISAC 112 validation ; ISAC 112 reliability </wiki> \newline
       <graph> ISAC 112 , reproducibility ; specific IgE , component-resolved diagnostics ; allergy testing , reliability </graph>
       \\
       \midrule 
      \bf SeRTS & 
       <book> reproducibility of ISAC 112 ; specific IgE component testing reproducibility ; validation of ISAC 112 technique </book> \newline
       \textcolor{purple}{<guideline> reproducibility of ISAC 112 ; specific IgE component testing ; recommendations for ISAC 112 use </guideline>} \newline
       <research> reproducibility studies of ISAC 112 ; validation of specific IgE component testing </research> \newline
       <wiki> ISAC 112 reproducibility ; specific IgE component testing </wiki> \newline
       <graph> component-resolved diagnostics , ISAC 112 ; specific IgE , component-resolved diagnostics ; allergy testing , reliability </graph>
       \\
       \midrule
       \bf SPO Planning &
       <book> reproducibility of ISAC 112 in specific IgE determination ; ISAC 112 and specific IgE testing </book> \newline
       \textcolor{mydrakgreen}{<guideline> </guideline>} \newline
       \textcolor{mydrakgreen}{<research> ISAC 112 accuracy in specific IgE determination ; ISAC 112 reliability in component-resolved diagnostics ; ISAC 112 in allergy research </research>} \newline
       <wiki> ISAC 112 and specific IgE testing </wiki> \newline
       <graph> </graph>
       \\
       \bottomrule[1pt]
    \end{tabular}
}
\caption{Two case studies comparing the different source planning methods.}
\label{tab:case_study}
\end{table*}

Two case studies are conducted as shown in Table~\ref{tab:case_study}. In the first case, it is difficult to retrieve useful information from various knowledge sources with such a complex medical diagnosis as a search term. Our proposed SPO is able to propose queries that better fit the characteristics of the sources, compared to Prompting and SeRTS. Specifically, appropriate queries for the ``Guideline'' source are constructed by SPO, whereas queries proposed by Prompting and SeRTS are only roughly descriptive and have a high degree of duplication with other sources' queries. Moreover, for such complex problems, queries with more perspectives are generated by SPO, such as those in the ``Wiki'', to provide comprehensive knowledge. In the second case, where a concise biomedical question is presented, the SPO method prioritises the most relevant ``Research'' source. In contrast, the other two methods continue to consider the irrelevant ``Guideline'' source, primarily due to their limited awareness of the multi-source knowledge base.

\section{Additional Details}

\subsection{MedOmniKB Details}\label{appendix:db}

The category statistics of gathered books are detailed in Table~\ref{tab:stat_books}.

For datasets that provide reference documents (MedQA, PubMedQA, BioASQ, LiveQA, MedicationQA), MedOmniKB includes either the original reference materials or similar online resources. For datasets that do not explicitly provide reference documents (MedMCQA, MMLU-Med, SEER, DDXPlus, MIMIC-IV-ED, ExpertQA-Biomed), MedOmniKB is expected to meet knowledge requirements through its extensive and diverse knowledge entries.

Importantly, this scenario simulates real-world medical scenarios, where annotated reference documents for health issues are often unavailable. The superior SPO performances on these datasets, which lack reference annotations, further demonstrate its reliability and potential for broad medical applications.

\subsection{Dataset Details}\label{appendix:train_test_data}

\begin{table*}[htbp]
\centering
    \resizebox{\linewidth}{!}{
    \begin{tabular}{lcc} 
        \toprule[1pt]
        \bf Dataset & \bf Type & \bf Medical Application \\
        \midrule
MedQA & \multirow{3}{*}{Reasoning QA} & Medical Reasoning Involving Diagnoses, Treatments, and Examinations \\
MedMCQA & & Commonsense Reasoning Focused on Medical Knowledge \\
MMLU-Med & & Commonsense Reasoning Focused on Medical Knowledge \\
\midrule
PubMedQA & \multirow{2}{*}{Research QA} & Knowledge Acquisition Centered on Advanced Research \\
BioASQ & & Knowledge Acquisition Centered on Advanced Research \\
\midrule
SEER & \multirow{3}{*}{Clinical QA} & Treatment Planning for Patients with Different Disease Severity \\
DDXPlus & & Diagnostic Based on Patient Dialogue Records \\
MIMIC-IV-ED & & Clinical Outcome Prediction Using EHR Records for Decision-Making \\
\midrule
LiveQA & \multirow{3}{*}{Long-form Answering} & Consumer Health Inquiry Support Across Various Domains \\
MedicationQA & & Consumer Health Inquiry Support Regarding Medications \\
ExpertQA-Biomed & & Professional-Level Biology and Healthcare Solutions \\

        \bottomrule[1pt]
    \end{tabular}
    }
    \caption{The type and medical application of each dataset.}
    \label{tab:dataset_app}
\end{table*}

The 11 datasets adopted in our work are shown in Table~\ref{tab:dataset_stat} and Table~\ref{tab:dataset_app}.
\begin{itemize}
 
\item \textbf{MedQA}~\cite{Jin2020} includes clinical medicine questions from the United States Medical Licensing Examination, covering topics such as diagnoses, treatments, and examinations. We adopt their 4-option English subset. In particular, 6000 training samples, 1200 development samples and 1273 test samples are selected from their official sets.

\item \textbf{MedMCQA}~\cite{Pal2022} are collected from Indian medical entrance exams. Their questions cover 2400 healthcare topics and 21 medical subjects with high topical diversity. We randomly sample 6000 training samples from the official training split. The 1200 development samples and 1200 test samples are randomly selected from the official development set since their test set lacks labelled answers.

\item \textbf{MMLU-Med}~\cite{Hendrycks2021} comprises six tasks related to biomedicine, which include anatomy, college biology, college medicine, clinical knowledge, human genetics and professional medicine. There are a total of 1089 test samples. MMLU-Med is included solely in our test set due to the limited number of training and development samples available in the dataset.

\item \textbf{PubMedQA}~\cite{Jin2019} is a biomedical QA dataset, containing 1000 manually annotated questions based on PubMed abstracts. We remove the origin-given contexts for the RAG setting. The original 500 test samples are adopted as our test set. For the remaining samples, 417 samples are included in the training set, while 83 samples are included in the development set.

\item \textbf{BioASQ}~\cite{Tsatsaronis2015} is from an annual competition for biomedical question-answering\footnote{\url{https://www.bioasq.org/}}. Following \citet{Xiong2024}, we select the Yes/No questions in Task B of the competition from 2014 to 2024. The questions are also based on biomedical literature. We remove the given context for the RAG setting. We randomly divide all available samples into three portions: 584 samples for the training set, 116 samples for the development set, and 782 samples for the test set.

\item \textbf{SEER}~\cite{Dubey2023} acts as a treatment planning dataset involving the Surveillance, Epidemiology, and End Results (SEER) custom breast cancer databases. For each patient record, 17 attributes and recommended treatment plans are noted. We borrow the processed dataset from MedS-Bench~\cite{Wu2025}, which simplifies the task into a 7-option recommendation format. We randomly divide samples into three portions: 2113 samples for the training set, 422 samples for the development set, and 1000 samples for the test set.

\item \textbf{DDXPlus}~\cite{Tchango2022} is an advanced large-scale dataset designed for Automatic Symptom Detection (ASD) and Automatic Diagnosis (AD) systems. LLMs must make diagnostic decisions based on dialogues and select from 49 potential diagnoses. We also adopt the data from MedS-Bench~\cite{Wu2025}. We randomly divide samples into three portions: 5000 samples for the training set, 1000 samples for the development set, and 1000 samples for the test set.

\item \textbf{MIMIC-IV-ED}~\cite{Xie2022} is a standardised benchmark derived from the Medical Information Mart for Intensive Care IV-Emergency Department (MIMIC-IV-ED) database. Following MedS-Bench~\cite{Wu2025}, three medical tasks are performed. Given a patient's EHR, LLMs are required to predict whether the patient needs hospitalisation, needs to revisit the emergency department within 72 hours, or should be classified into a critical triage queue. They can help clinicians make clinical decisions. We randomly divide samples into three portions: 2917 samples for the training set, 583 samples for the development set, and 1000 samples for the test set.

\item \textbf{LiveQA}~\cite{Abacha2017} focus on answering consumer health questions received by the U.S. National Library of Medicine. Manually collected and validated reference answers from trusted sources, such as the National Institute, are provided. We adopt the 104 original test samples as the test set. The remaining samples are divided into two portions: 343 samples for the training set and 68 samples for the development set.

\item \textbf{MedicationQA}~\cite{Abacha2019} is constructed by collecting anonymized consumer questions submitted to MedlinePlus\footnote{\url{https://medlineplus.gov/}}. Four annotators participated in the manual annotation and answering process. Finally, the answers are validated by medical experts. We randomly divide samples into three portions: 450 samples for the training set, 90 samples for the development set, and 150 samples for the test set.

\item \textbf{ExpertQA-Biomed}~\cite{Malaviya2024} is a high-quality long-form QA dataset across 32 fields. The dataset comprises questions and answers proposed and verified by domain experts. Among them, 96 biology (Bio) questions and 504 medical questions (Med) are used in this work. We randomly divide these samples into three portions: 375 samples for the training set, 75 samples for the development set, and 150 samples for the test set.

\end{itemize}

\subsection{Baseline Details}\label{appendix:baselines}

\begin{itemize}
\item \textbf{No Retrieval}: The reader is prompted to think step-by-step and answer the question without external documents (Prompt~\ref{prompt:reader_nodoc}).

\item \textbf{Original Question}~\cite{Xiong2024}: The MedRAG method, which uses the original question to retrieve a single mixed source, is adapted to our multi-source setting.
Specifically, the original question is used as the query for the four unstructured sources: ``Book'', ``Guideline'', ``Research'' and ``Wiki''. For the ``Graph'' source, we prompt the Qwen2.5-72B-Instruct-AWQ to extract medical terms of less than 4 (Prompt~\ref{prompt:extract_terms}), which will be used as queried terms.

\item \textbf{Query2Doc}~\cite{Wang2023}: We prompt the Qwen2.5-72B-Instruct-AWQ to enhance original question by producing pseudo-documents. For the four unstructured sources, we produce the documents in the corresponding style (Prompt~\ref{prompt:query2doc}). For the structured ``Graph'' source, we use the same approach as in the Original Question.

\item \textbf{Prompting}~\cite{Ma2023}: Qwen2.5-72B-Instruct-AWQ is directly prompted to construct proper queries for each source. The number of queries for each source is limited to 0-3.

\item \textbf{Reflexion}~\cite{Shinn2023}: Self-reflection on a single reasoning path is conducted in this baseline. Following \citet{Hu2024}, we iteratively carry out the following steps: 1. Construct appropriate queries for each source by incorporating all previous feedback (if any) (Prompt~\ref{prompt:reflect}). 2. Generate feedback and assign scores for the new plan (Prompt~\ref{prompt:reflect_feedback_score}). If the score reaches 5 (the maximum), the loop stops. The maximum number of iterations is set to 8 due to significant inference costs. This method is adapted to our multi-source setting by proposing and evaluating queries for all sources.

\item \textbf{SeRTS}~\cite{Hu2024}: Based on Reflexion, sibling nodes are added and join the reasoning process. We follow \citet{Hu2024} to perform the four operations (Selection, Expansion, Evaluation and Backpropagation) iteratively. If the score reaches 5 (the maximum), the loop stops. The maximum number of iterations is set to 8 due to significant inference costs. This method is adapted to our multi-source setting by proposing and evaluating queries for each source.

\item \textbf{Trainable Planning}~\cite{Ma2023}: Different from the proposed SPO Planning, we use the performance of Qwen2.5-72B-Instruct-AWQ on the downstream tasks to measure the value of each query. The metric is the improvement of accuracy for closed-ended QA, and Rouge-L~\cite{Lin2004} for open-ended generation. The other SFT and DPO training processes remain the same as in SPO Planning.

\item \textbf{RaFe Planning}~\cite{Mao2024}: Unlike the proposed SPO Planning, we use the rerank scores between the retrieved documents and the question to assess the value of each query. The rerank score threshold is set to 2.3 by trying on the development set. The other SFT and DPO training processes remain the same as in SPO Planning.

\end{itemize}

For the baselines involving retrieval, the reader will read the retrieved documents and finally answer the question after step-by-step thinking (Prompt~\ref{prompt:reader_doc}).

\subsection{Implementation Details}\label{appendix:imple}

We employ the superior LLM, Qwen2.5-72B-Instruct~\cite{QwenTeam2024} to explore multiple plans and verify their correctness. The Activation-aware Weight Quantization (AWQ)~\cite{Lin2024} accelerates inference and reduces memory requirements.

We adopt Qwen2.5-7B-Instruct as the backbone of the SPO process. Low-Rank Adaptation (LoRA)~\cite{Hu2022} is adopted in the SFT and DPO process for Parameter-Efficient Fine-Tuning. For the hyperparameters of the SFT process, the batch size is 64, the peak learning rate is 2.5e-4, and the number of epochs is 5. As for the DPO process, the batch size is 64, the peak learning rate is 5e-6, and the number of epochs is 3. We use vLLM~\cite{Kwon2023} for fast inference and set the temperature to 0 for reproducible outcomes.

\section{Prompt List}\label{appendix:prompt}

\input{08a_prompts}

%% file: 08a_prompts.tex
\begin{figure*}[htbp]
\begin{prompt}[title={Prompt \thetcbcounter: Description of sources in MedOmniKB}, label=prompt:source_desc]
book: The API provides access to medical knowledge resources, including various educational resources and textbooks.\\
\\
guideline: The API provides access to clinical guidelines from leading health organizations.\\
\\
research: The API provides access to advanced biomedical research, facilitating access to specialized knowledge and resources.\\
\\
wiki: The API provides access to general knowledge across a wide range of topics.\\
\\
graph: The API provides a structured knowledge graph that connects medical definitions and related terms.
\end{prompt}
\end{figure*}

\begin{figure*}[htbp]
\begin{prompt}[title={Prompt \thetcbcounter: Query formats of sources in MedOmniKB}, label=prompt:source_format]
book: \{search\_query0\} ; \{search\_query1\} ; ... (Use ; to separate the queries, 0 to 3 queries)\\
\\
guideline: \{search\_query0\} ; \{search\_query1\} ; ... (Use ; to separate the queries, 0 to 3 queries)\\
\\
research: \{search\_query0\} ; \{search\_query1\} ; ... (Use ; to separate the queries, 0 to 3 queries)\\
\\
wiki: \{search\_query0\} ; \{search\_query1\} ; ... (Use ; to separate the queries, 0 to 3 queries)\\
\\
graph: \{medical\_term0\} , \{query\_for\_term0\} ; \{medical\_term1\} , \{query\_for\_term1\} ; ... (Use ; to separate the queries, 0 to 3 queries. Each query should use , to separate the \{medical\_term\} and \{query\_for\_term\})
\end{prompt}
\end{figure*}

\begin{figure*}[htbp]
\begin{prompt}[title={Prompt \thetcbcounter: Planning exploration}, label=prompt:plan_explor]
To answer the question labeled as \# Question, please construct appropriate queries to get the information you need.\\
\\
1. Each source in \# Source Description must have search queries constructed.\\
2. Please give the search queries following the format in \# Query Format. Each source should have \{n\_queries\} queries, separated by `;`. Please ensure the diversity of queries from the same source.\\
3. The queries for the source should accurately reflect the specific information needs from that source.\\
\\
\# Question\\
\{question\}\\
\\
\# Source Description\\
(Prompt~\ref{prompt:source_desc}, ``0 to 3 queries'' is removed)\\
\\
\# Query Format\\
(Prompt~\ref{prompt:source_format}, ``0 to 3 queries'' is removed)
\end{prompt}
\end{figure*}

\begin{figure*}[htbp]
\begin{prompt}[title={Prompt \thetcbcounter: Planning judging}, label=prompt:plan_judge]
You are a professional medical expert. Please judge whether the information in the \# Documents supports the \# Gold Answer as a response to the \# Question. Please first think step-by-step and then show your judgement. Your responses will be used for research purposes only, so please have a definite answer.\\
You should respond to the following question using the format <answer>yes/no</answer> at the end of your response. Please keep your entire response simple and complete, up to 100 words.\\
\\
\# Question\\
\{question\}\\
\\
\# Gold Answer\\
\{gold\}\\
\\
\# Documents\\
\{documents\}\\
\\
Hint: Please judge whether \# Documents supports the \# Gold Answer in response to the \# Question, rather than evaluating if the \# Question's answer is the \# Gold Answer.
\end{prompt}
\end{figure*}

\begin{figure*}[htbp]
\begin{prompt}[title={Prompt \thetcbcounter: Original Question (extract terms for querying Graph)}, label=prompt:extract_terms]
You are a helpful medical expert. Please return all medical terminologies in the input \# Question. Each term should be split by `,`.\\
The output format should be like:\\
<term>term0 , term1 , ...</term> (The number of terms should be less than 4)\\
\\
\# Question\\
\{question\}
\end{prompt}
\end{figure*}

\begin{figure*}[htbp]
\begin{prompt}[title={Prompt \thetcbcounter: Query2Doc}, label=prompt:query2doc]
You are a professional medical expert. Please write a passage that answers the given \# Question. The passage should be considered as part of the source described in the \# Source Description provided. The result should be formatted as <passage>your passage</passage>. Please keep your entire passage up to 200 words.\\
\\
\# Question\\
\{question\}\\
\\
\# Source Description\\
\{source\_name\}: \{source\_desc\} (For a given source, provide its source name and description, which can be found in Prompt~\ref{prompt:source_desc}.)
\end{prompt}
\end{figure*}

\begin{figure*}[htbp]
\begin{prompt}[title={Prompt \thetcbcounter: Prompting, Trainable Planning, RaFe Planning, SPO Planning}, label=prompt:plan]
To answer the question labeled as \# Question, please construct appropriate queries to get the information you need.\\
\\
1. Please give the search queries following the format in \# Query Format. The source can have up to 3 queries, separated by `;`. Please ensure the diversity of queries from the same source. For each source, if you think no information retrieval is needed, simply output an empty tag for that source, for example: <book></book>.\\
2. The queries for the source should accurately reflect the specific information needs from that source.\\
\\
\# Question\\
\{question\}\\
\\
\# Source Description\\
(Prompt~\ref{prompt:source_desc})\\
\\
\# Query Format\\
(Prompt~\ref{prompt:source_format})
\end{prompt}
\end{figure*}

\begin{figure*}[htbp]
\begin{prompt}[title={Prompt \thetcbcounter: Reflextion and SeRTS (reflect and propose improved queries)}, label=prompt:reflect]
To answer the question labeled as \# Question, please construct appropriate queries to get the information you need. Pay attention to \# Feedback from previous search queries.\\
1. Please give the search queries following the format in \# Query Format. The source can have up to 3 queries, separated by `;`. Please ensure the diversity of queries from the same source. For each source, if you think no information retrieval is needed, simply output an empty tag for that source, for example: <book></book>.\\
2. The queries for the source should accurately reflect the specific information needs from that source.\\
\\
\# Question\\
\{question\}\\
\\
\# Feedback\\
\{feedback\}\\
\\
\# Source Description\\
(Prompt~\ref{prompt:source_desc})\\
\\
\# Query Format\\
(Prompt~\ref{prompt:source_format})\\
\\
Please review the multiple queries and corresponding suggestions in \# Feedback, and construct improved queries.
\end{prompt}
\end{figure*}

\begin{figure*}[htbp]
\begin{prompt}[title={Prompt \thetcbcounter: Reflextion and SeRTS (produce the feedback and score)}, label=prompt:reflect_feedback_score]
You are a highly intelligent agent. I am currently answering the \# Question. Then, I have retrieved relevant content \# Documents using corresponding \#\# source: xxx; query: xxx. \\
1. Please rate \# Documents using the additive 5-point scoring system described below. Points are accumulated based on the satisfaction of each criterion:\\
\{Five-point Rubrics\}\\
2. Please give suggestions for constructing better queries for each source. You should also consider not using certain sources.\\
\\
\# Question\\
\{question\}\\
\\
\# Documents\\
\{docs\}\\
\\
\# Source Description\\
\{source\_desc\}\\
\\
Please provide the suggestion using the <suggestion></suggestion> tags. You should consider enabling unused sources or abandoning used ones. Please keep the suggestion within 100 words.\\
Please score between 0 and 5, strictly using the aforementioned additive 5-point scoring system and the format: <score> Integer Score </score>.\\
\\
<suggestion>Your suggestion here</suggestion>\\
<score>Your score here</score>
\end{prompt}
\end{figure*}

\begin{figure*}[htbp]
\begin{prompt}[title={Prompt \thetcbcounter: Reader without retrieved documents}, label=prompt:reader_nodoc]
You are a professional medical expert to answer the \# Question. Please first think step-by-step and then answer the question. Your responses will be used for research purposes only, so please have a definite answer.\\
You should think step by step and respond in the format <answer>A/B/C/...</answer> (only one option can be chosen) at the end of your response. Please keep your entire response simple and complete, up to 100 words. (Option restrictions are removed in open-ended generation.)\\
\\
\# Question\\
\{question\}
\end{prompt}
\end{figure*}

\begin{figure*}[htbp]
\begin{prompt}[title={Prompt \thetcbcounter: Reader with retrieved documents}, label=prompt:reader_doc]
You are a professional medical expert to answer the \# Question. Please first think step-by-step using the \# Retrieved Documents and then answer the question. Your responses will be used for research purposes only, so please have a definite answer.\\
You should think step by step and respond in the format <answer>A/B/C/...</answer> (only one option can be chosen) at the end of your response. Please keep your entire response simple and complete, up to 100 words. (Option restrictions are removed in open-ended generation.)\\
\\
\# Retrieved Documents\\
\{documents\}\\
\\
\# Question\\
\{question\}
\end{prompt}
\end{figure*}

\begin{figure*}[htbp]
\begin{prompt}[title={Instruction \thetcbcounter: Instruction for human evaluation for long-form answering}, label=prompt:human_rate]
Relevance:\\
1: Completely unrelated to the question\\
2: Some relation to the question, but mostly off-topic\\
3: Relevant, but lacking focus or key details\\
4: Highly relevant, addressing the main aspects of the question\\
5: Directly relevant and precisely targeted to the question\\
\\
Completeness:\\
1: Extremely incomplete\\
2: Almost incomplete with limited information\\
3: Moderate completeness with some information\\
4: Mostly complete with most of the information displayed\\
5: Fully complete with all information presented\\
\\
Proficiency in medicine:\\
1: Using plain language with no medical terminology.\\
2: Equipped with some medical knowledge but lacking in-depth details\\
3: Conveying moderately complex medical information with clarity\\
4: Showing solid grasp of medical terminology but having some minor mistakes in detail \\
5: Fully correct in all presented medical knowledge\\
\\
Interpretability (focused on retrieved documents):\\
1: Minimal relevant information; does not assist in answering the question\\
2: Limited relevant content; offers little help in answering the question\\
3: Some relevant information; may need further clarification to answer the question\\
4: Most relevant information presented; generally aids in answering the question but lacks some details\\
5: Comprehensive and clear; fully supports answering the question\\
\\
\# Question\\
\{question\}\\
\\
\# Reference Answer\\
\{reference\}\\
\\
\# Predicted Answer \\
\{prediction\}
\end{prompt}
\end{figure*}